\documentclass[10pt,twocolumn,letterpaper]{article}

\usepackage{cvpr}
\usepackage{times}
\usepackage{epsfig}
\usepackage{graphicx}
\usepackage{amsmath}
\usepackage{amssymb}
\usepackage{diagbox}
\usepackage{tabularx}
\usepackage{gensymb}
\usepackage{makecell}
\usepackage{tabularx}

\usepackage[breaklinks=true,bookmarks=false]{hyperref}

\cvprfinalcopy 

\begin{document}


\title{Pose-based Deep Gait Recognition}

\author{Anna Sokolova$^1$, Anton Konushin$^{1,2}$\\
\small 1. National Research University Higher School of Economics, 20 Myasnitskaya str., 
Moscow 101000, Russia\\
\small 2. Lomonosov Moscow State University, GSP-1, Leninskie Gory, Moscow, 119991, Russia}

\maketitle

\begin{abstract}
Human gait or walking manner is a biometric feature that allows identification of a person when other biometric features such as the face or iris are not visible. In this paper, we present a new pose-based convolutional neural network model for gait recognition. Unlike many methods that consider the full-height silhouette of a moving person, we consider the motion of points in the areas around human joints. To extract motion information, we estimate the optical flow between consecutive frames. We propose a deep convolutional model that computes pose-based gait descriptors. We compare different network architectures and aggregation methods and experimentally assess various sets of body parts to determine which are the most important for gait recognition. In addition, we investigate the generalization ability of the developed algorithms by transferring them between datasets. The results of these experiments show that our approach outperforms state-of-the-art methods.
\end{abstract}

\section{Introduction}

Gait recognition is a computer vision problem that comprises the identification of an individual in a video using the motion of their body as the only source of information. 
Unlike face recognition or re-identification problems,  gait recognition does not rely solely on an individual's appearance, which tends to make the task more complex. Physiological studies show that each person has his/her own unique manner of walking, which is difficult to forge; thus, a person can be identified by the gait.

Gait recognition methods have several advantages that make them usable in many applied problems. First, unlike the face, iris, or fingerprints, gait representation can be recorded without a subject's cooperation. The second advantage is that motion can be captured and a person can be recognized even from a video with low resolution. Such features are very significant in video surveillance, and thus, gait recognition has become an important field of inquiry, particularly in the security field, the primary goals of which are the control of access to restricted areas and the detection of people who have previously been captured on camera (e.g., criminals).

Despite the uniqueness of gait, there are many factors that can affect gait representation and make the problem more complicated. Gait can appear to differ depending on the viewing angle or the clothing worn by the subject. Furthermore, wearing different shoes or carrying heavy bags will change the gait itself, and a recognition algorithm should be stable to such changes. 

The problem of gait recognition is closely related to several computer vision problems. First, it is an identification problem similar to face recognition, with the difference that in gait recognition the motion of the body, but not the appearance, is of importance. A person may wear a mask veiling the face or a coat hiding the figure, but the body motion will still be the same.  In addition, as well as action recognition it is a video classification problem, thus, the gait recognition problem may be solved by the same methods as action recognition.
A third problem approximated to gait recognition  is re-identification (re-id); however, as in face recognition re-id addresses appearance rather than motion.

\begin{figure*}[ht]
\begin{center}
		\includegraphics[width=0.9\linewidth]{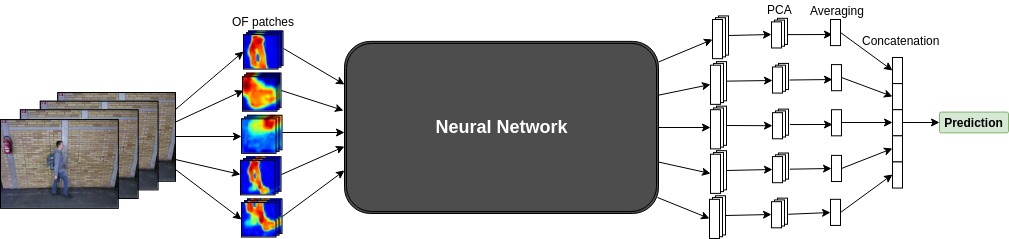}
	\caption{The pipeline of the algorithm.}
\label{fig1}
\end{center}
\end{figure*}

The similarity of these problems means that we can use approaches from adjacent fields in gait recognition. Most modern computer vision methods are based on convolutional neural networks (CNNs) and can be transformed for gait recognition. However, despite the development of such algorithms the most successful gait recognition approaches are still not deep and use handcrafted features. In this work, we propose a CNN-based algorithm for recognition of people based on their walking manner. This method proves to be more stable in transfer learning and achieves higher classification accuracy than previous deep gait models.

Because we investigate the motion of the body, the person's appearance should not be taken into the account. Accordingly, we can consider the optical flow as the main source of information and avoid the use of raw images. Our experiments reveal that such an approach does obtain sufficient data and produces successful results.

\section{Related Work}

Currently, there are two leading approaches to gait recognition. The more traditional approach is based on the extraction of handcrafted features from image frames. Most investigations following this approach use silhouette masks as the main source of information and extract features that show how these masks change. The most popular descriptor of gait used in such investigations is the gait energy image (GEI) \cite{han}, a binary mask averaged over the gait cycle of a human figure. This approach has developed significantly during recent years. Many different descriptors have been proposed for application to GEI (e.g., the HOG \cite{liu} and HOF descriptors) or to entire silhouette sequences, e.g., the frame difference energy image (FDEI) \cite{fdei} to provide additional aggregation for better gait representation.  GEIs are also used in a unified metric learning framework \cite{joint_int} where joint intensity and spacial metric are optimized in order to mitigate the intrasubject differences and increase the intersubject ones.

One more gait representation technique based on silhouette extraction is frequency-domain gait entropy (EnDFT) \cite{EnDFT}. This approach combines two other techniques: discrete fourier transform (DFT) \cite{DFT} and gait entropy image (GEnI) \cite{GEnI}. In the first one the mean of binary silhouette masks with exponentially decreasing weights is calculated, while the second one computes the entropy of each point of silhouette over the gait period. GEnI is the mixture of these approaches, the entropy is computed from DFT instead of GEI. Besides, the most effective human body parts are used for the recognition according to the recognition accuracy of the rows of DFT gait features.

The highest cross-view recognition quality is shown by Li \textit{et al.} \cite{bayes} who proposed the bayesian approach. GEI is supposed to be a sum of gait identity value and some noise standing for different gait variations (view, clothing and carrying bags etc.) both following Gaussian distribution. Covariance matrices of these distributions are optimized by EM algorithm considering joint distribution of GEI pairs.

Another approach to gait recognition is neural networks. Deep models have been used to obtain the best results in most computer vision problems and have recently inspired new investigations of gait based on CNN. The similarity of gait and action recognition problems means that many approaches applied to the latter can be used for the former. The first and most classical deep method was proposed by Simonyan and Zisserman \cite{simonyan}. To recognize human actions, they trained a network with an architecture comprising two similar types of branches: image and flow streams. The former processes raw frames of video, while the latter gets the maps of optical flow (OF) computed from pairs of consecutive frames as input. To consider long-duration actions, several consecutive flow maps are stacked into blocks that are used as network inputs.
Many action recognition algorithms are based on this framework apply variant top architectures (including the recurrent architectures \cite{beyond} and methods for fusing several streams \cite{Feichtenhofer}).
Another method based on optical flow considers temporal information using three-dimensional convolutions \cite{varol17}.

The most applicable approach to gait recognition is P-CNN \cite{pcnn}. Similar to previous methods, this method used two streams but, in place of full-body maps, each stream receives patches in which different body parts are cropped. Thus, some joints are considered more precisely, which helps in the recording of small but important bodily motions.

The OF approach was applied to the gait recognition problem in several works \cite{castro,me}, which proposed a deep model that uses blocks of OF maps containing full bodies as inputs to predict recorded individuals. 
Several other deep gait recognition solutions unite neural and GEI approaches. GEINet \cite{geinet} computes gait energy images at various viewing angles for network input. In DeepGait \cite{deep_gait} neural network features are extracted using silhouette masks as input and maximum response over the gait cycle is used.
Wu \textit{et al.} \cite{cross_view_dcnn} proposed a deep CNN that predicts the similarity given a pair of gait sequences and considered different ways of comparing gait features.
However, despite the success of neural network approaches such as these, many non-deep methods still achieve a higher quality of gait recognition.

\section{Proposed Method}\label{Proposed Method}
Here, we describe the pipeline of our proposed method. Although the algorithm is based on neural networks, two important data pre-processing stages are applied: motion map computation and frame-by-frame evaluation of the individual's pose. After these steps are completed, the network can be trained to classify video sequences.

Let us discuss all the stages of the algorithm.
\subsection{Preprocessing the data}\label{Preprocessing the data}

As our goal is to train a feature extractor that does not depend on, for example, clothing color or personal appearance, we can eliminate all color information and use only motion. To do this, we compute maps of optical flow between each pair of consecutive frames and treat these maps as inputs. We consider three-channel OF maps in which the first two channels carry, respectively, horizontal and vertical components of flow vectors and the third carries magnitude. Prior to further processing, all maps are linearly transformed to the interval $[0,255]$ in a coding similar to that used in RGB channels.

\begin{figure}[ht]
\begin{center}
		\includegraphics[width=.7\columnwidth]{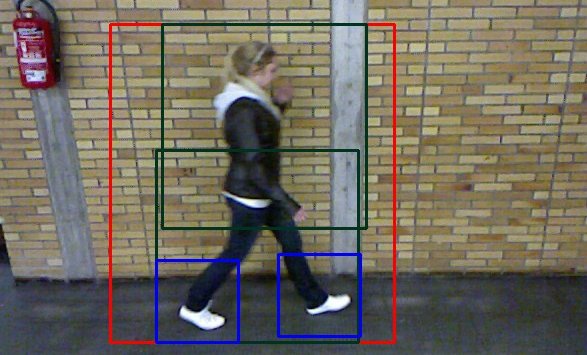}
	\caption{The bounding boxes for human body parts: right and left feet (blue boxes), upper and lower body (green boxes) and full body (red box).}
\label{fig2}
\end{center}
\end{figure}

We can further assume that the motions of some parts of the human body are more informative than others and, correspondingly, evaluate the human pose and limit our analysis of optical flow maps to the neighborhoods of such key positions. We would expect that the bottom part of the body carries more gait information and therefore pay more attention to the legs than to either the hands or the head. We therefore estimate the pose of the human in each frame (by finding joint locations) and crop five patches from the obtained OF maps: right foot, left foot, upper body, lower body, and full body. Bounding boxes of these body parts are shown in Fig.~\ref{fig2}.

The leg patches are squares with key leg points in their centers, the upper body patch contains all of the joints from the head to the hips (including the hands), and the lower body patch contains all of the joints from the hips to the feet (excluding the hands). Thus, each pair of consecutive frames produces five patches for use as network inputs. Prior to inputting the data into the network, the resolution of each patch is decreased to $48\times48$ pixels.

\subsection{Data augmentation}\label{Data augmentation}
The primary component of the proposed algorithm involves the extraction of neural features. As every deep neural network has a large number of parameters, it is necessary to augment the data to obtain a stable and not overfitted algorithm. 

The data are augmented using classic spatial changes. In the training process, four numbers are uniformly sampled for each of the five considered body parts to construct the bounds of an input patch. The first two numbers are left and right extensions chosen from the interval $[0,w/3]$, where $w$ is the width of the initial bounding box, while the latter two numbers are upper and lower extensions in $[0,h/3]$, where $h$ is the height of the box.
The resulting bounded patches are cropped from the OF maps and then resized to $48\times48$.

This augmentation allows us to obtain patches containing body parts undergoing both spatial shifts and zoom. If the sum of the sampled numbers is close to zero, a large image of the body part with very little excess background is obtained; otherwise, the image is smaller with more background. On the other hand, fixing the sums of the first and the second pairs of bounds produces body parts with the same size but changed location inside the patch.

\subsection{Training the neural network}\label{Training the neural network}

The network is trained using the augmented data produced above and then used as a feature extractor, with the outputs of the last hidden layer used as gait descriptors. In the testing stage, instead of sampling random bounds for each patch, their mean value is taken ($w/6$ and $h/6$, respectively) to locate the body part in the center of the patch.

\subsection*{CNN architectures and training methods}
We considered and compared two network architectures. The first is based on the VGG-19 \cite{vgg} network but has one less convolutional block; details of this architecture are shown in Table~\ref{tab1}. 
\begin{table}[h]
\caption{The VGG-like architecture.\label{tab1}}
	{\centering
		\begin{tabular}{|p{0.1\linewidth}|p{0.12\linewidth}|p{0.12\linewidth}|p{0.12\linewidth}|p{0.07\linewidth}|p{0.07\linewidth}|c|c|}\hline
			\multicolumn{1}{|c|}{B1}&\multicolumn{1}{|c|}{B2}&\multicolumn{1}{|c|}{B3}&\multicolumn{1}{|c|}{B4}&\multicolumn{1}{|c|}{F5}&\multicolumn{1}{|c|}{F6}&SM\\\hline
			  
			 3x3,64& 3x3,128 &3x3,256 &3x3,512 &   & & \\
			 3x3,64& 3x3,128 &3x3,256 &3x3,512 & 4096 &4096&soft- \\
              &   &3x3,256 & 3x3,512 & d/o& d/o& max\\
			 pool~2& pool 2& pool 2 &pool 2 &&  & \\\hline
		\end{tabular}}{}
\end{table}

Each column of this table corresponds to a block of layers. The first four blocks are convolutions, with each row denoting, for each layer in the block, the size of its filters ($3\times3$ for all the layers) and the number of filters; the number of filters per block is doubled in each succeeding block. Additionally, there are four max-pooling layers of size $2\times2$ after each convolutional block.

The next two blocks are fully connected, with each comprising one linear dense layer of size 4,096 and a dropout (d/o) with the probability parameter $p=0.5$. In addition to convolutions, the dense layers are followed by ReLU non-linearities. The final column denotes the top block, comprising a dense layer and a softmax non-linearity. The number of units in this block is equal to the number of training subjects that can train the network for a classification task.

During training, an $L_2$ norm of dense layer weights is added to the loss function to enhance the regularization against overfitting.

Networks similar to this have produced the best results in previous experiments \cite{me} in which the blocks of several consecutive OF maps were used as network inputs. However, that approach \cite{me} did not take into account key points of the human pose; instead, the full body was used. We will further compare this previous study with our approach in Section 4.

Similar to the block-based model \cite{me}, we trained our network step by step. We started from 1,024 units in two hidden dense layers and trained the network while doubling the layer size each time the accuracy ceased to increase. When the layer size reached 4,096 units, the training stopped. Each widening of the network was implemented by random initialization of new parameters to add extra regularization and prevent overfitting while training.

After developing this deep CNN with both convolutional and dense layers, we considered implementing a fully convolutional network. One of the more successful architectures in computer vision used for image classification task is the ResNet architecture, in which residual connectivity allows for the transfer of information from low to high levels and the addition of each new block increases the accuracy of classification. Furthermore, the absence of fully connected layers in the ResNet architectures reduces the number of parameters of the model. These features have made ResNet a very popular architecture and inspired our investigation of its use here. Although residual networks achieve great results, they require large numbers of layers to significantly improve their performance, and each new block significantly lengthens the training process. Another problem with deep networks is exploding or vanishing gradients as a result of very long paths from the last to the first layer during backpropagation. To avoid these problems, we used the Wide Residual Network \cite{wideresnet} with decreased depth and increased residual block width; the resulting reduction in the number of layers made the training process much faster and allowed the network to be optimized with less regularization.

Table~\ref{tab2} shows the details of our Wide ResNet architecture.
\newcolumntype{L}[1]{>{\raggedright\arraybackslash}p{#1}}

\begin{table}[h]
\caption{The Wide Residual Network architecture}
	\centering
		\begin{tabular}{| l | c | c |c |}\hline
			\multicolumn{1}{|c|}{B1}&\multicolumn{1}{|c|}{B2}&\multicolumn{1}{|c|}{B3}&\multicolumn{1}{|c|}{B4}\\\hline
		\begin{tabular}{p{0.04\linewidth}}
			 
			      \small 3x3,16\\
			      \small BN 
			 \end{tabular}& $\small\left.
			 \begin{tabular}{p{0.04\linewidth}}
			      \small BN \\
			      \small 3x3,64  \\
			      \small BN \\
			      \small 3x3,64
			 \end{tabular} \ \ \ \  \right]\small {\times3}$ &
			 $\left.\begin{tabular}{p{0.04\linewidth}}
			      \small BN\\
			      \small 3x3,128 \  \\
			      \small BN \\
			      \small 3x3,128\ 
			 \end{tabular}\ \ \  \ \  \right]\small \times3$&
			 $\left.\begin{tabular}{p{0.04\linewidth}}
			      \small BN\\
			      \small 3x3,256  \\
			      \small BN \\
			      \small 3x3,256  
			 \end{tabular} \ \ \ \ \  \right]\small \times3 $\\
		     stride 1&stride 1 &  stride 2&  stride 2\\
			 \hline	  
			 
		\end{tabular}
\label{tab2}
\end{table}

Each column in the table defines a set of convolution blocks with the same number of filters. As in VGG-like architectures, all of the convolutions have kernels of size $3\times3$. The first layer has $16$ filters and is followed by batch normalization (BN) and then three residual blocks, each comprising two convolutional layers with 64 filters with normalization between them. This block construction is classical for Wide ResNet architectures, and all of our blocks have similar structures. In each successive group the blocks have twice as many filters as in the previous group. Each first convolutional layer in groups B3--B4 has a stride with parameter $2$; therefore, the tensor size begins at $48\times48$ pixels and decreases to $24\times24$ and then $12\times12$ in groups B3 and B4, respectively.


Following the residual blocks is a final process necessary to make the network useful for classification. This comprises one additional batch normalization layer, an average pooling that "flattens" the sequence of maps $12\times12$ obtained following the convolutions to one vector and one dense layer with softmax on the top. The number of units in this dense layer is equal to the number of subjects in the training set.
All of the activations except for the final softmax are rectified linear units that follow the batch normalization layers.


\subsection{Final classification}

The network is trained to predict one of the subjects from the training set with a patch cropped from its OF map. As our goal is to construct a feature extractor that can be used without retraining or fine-tuning for any testing set, we use the outputs of the final hidden layer as the gait descriptors and construct a new classifier for them. As we assume that the gait descriptors of a given person are spatially close to each other, we can use one of the simplest methods, the Nearest Neighbor (NN) classifier. We make a further $L_2$ normalizations of all gait features vectors prior to fitting and classifying, as many studies have shown that having a uniform length across all vectors improves the accuracy of NN classification.

Although most classical measures of distance between two vectors are Euclidean, we also consider the use of the Manhattan distance as a metric. Despite the fact that normalization is always performed relative to the $L_2$ distance, in most experiments it has been shown that finding the closest descriptor with respect to $L_1$ metrics produces better and more stable results.

To improve and speed up the classification, we reduce the dimensionality of the feature vectors using a principal component analysis (PCA) algorithm to reduce noise in the data and accelerate the fitting of classifiers.

\subsection*{Fusion of feature vectors}
The trained neural network and fitted NN classifier allow us to predict which subjects have patches with optical flow around one of the body parts. As we are starting analysis from an initial video sequence, if we consider $j$ body parts we obtain $j$ patches for each pair of consecutive frames and therefore $(N-1)j$ descriptors for the video, where $N$ is the number of frames in the sequence. If we consider the frames separately, we can obtain $(N-1)j$ answers per video; however, we require only one. We investigate two methods for constructing one feature vector from all network outputs. The first, a "naive" method, involves averaging all of the descriptors by calculating the mean feature vector over all frames and all body parts. This approach is naive in that the descriptors corresponding to different body parts have different natures; even if we compute them using one network with the same weights, it would be expected that averaging the vectors would mix the components into a disordered result. Surprisingly, the accuracy achieved using this approach is very high, as will be shown through comparison with another approach in the next section.

The second and more self-consistent approach is averaging the descriptors over only time. After doing so, $j$ mean descriptors corresponding to each to $j$ body parts are obtained and concatenated to produce one final feature vector.

\section{Data and experiments}\label{Data and experiments}
\subsection{Datasets}\label{Datasets}

We evaluated the methods described in the preceding section on three popular gait databases: the "TUM Gait from Audio, Image and Depth" (TUM-GAID) \cite{TUM} dataset; CASIA Gait Dataset B \cite{casia}; and the OU-ISIR Gait Database, Large Population Dataset \cite{ou_isir}.

\begin{figure}[ht]
\begin{center}
		\includegraphics[width=.9\columnwidth]{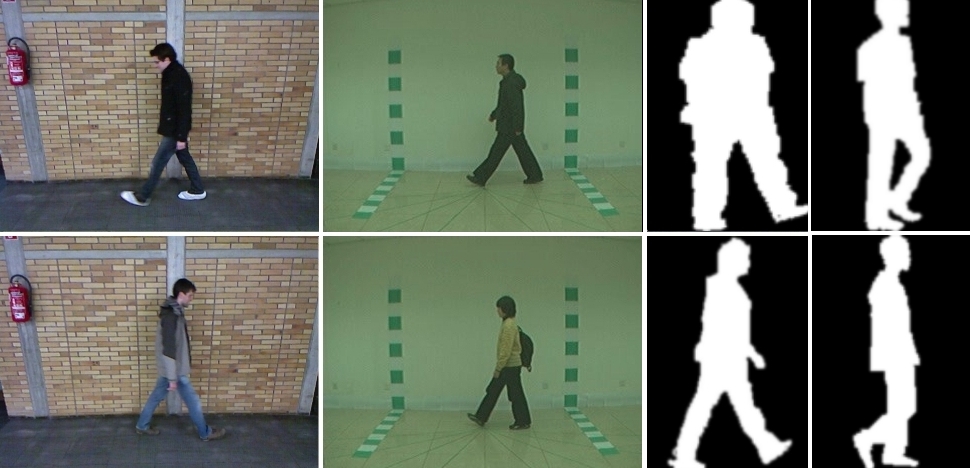}
	\caption{The examples of frames from two databases: TUM-GAID (the first column), CASIA Gait Dataset B (the second column) and OU-ISIR (the third and fourth columns).}
\label{fig3}
\end{center}
\end{figure}

The TUM-GAID dataset, comprising videos for $305$ subjects, is a sufficiently large database for gait recognition problems. The videos have lengths of 2--3 s apiece at a frame rate of 30 fps; we used all of them in our experiments. All videos are recordings of people of full height walking, captured from the side view. Although each person is recorded from only one viewpoint, there are several video sequences per subject taken under various conditions, e.g., wearing different shoes or carrying various items. Overall, there are 10 videos per subject: six normal walks, two walks in coating shoes, and two walks with a backpack. Examples of video frames are shown in Fig.~\ref{fig3} (first column). As is true of most gait databases, TUM-GAID's records contain one person per video without any intersection between figures. This is of course a naive approach, as in real life people often walk together with their bodies intersecting; nevertheless, this structure allows for the training of a model for checking if the problem of gait recognition can be theoretically solved. As the TUM-GAID is a relatively large database, it was the main data source in our experiments.

The second database we investigated was CASIA Gait Dataset B. This dataset contains only 124 subjects but is multiview: records are captured from 11 different viewpoints at angles ranging from 0 to 180 degrees. As in the TUM-GAID dataset, there are 10 videos per person captured in different conditions: normal walking, carrying a bag, and wearing a coat. Despite the fact that there are several sequences per subject and viewpoint, the dataset is very small for the purposes of deep modeling, as neural networks must contain many parameters, especially if they are to handle multiview modes. Thus, as we did for TUM-GAID we used only side-view videos captured from a constant angle and solved the side-view problem alone. The 124 subjects in this database represent a small set, even for the side-view mode; therefore, its use was limited to that of an additional database in some of the experiments. Frames from the CASIA database are shown in the second column of Fig.~\ref{fig3}.

The third dataset used was the OU-ISIR Gait Database. This is the largest gait database of the three, containing over 4,000 subjects, each captured by two cameras at four different angles (55, 65,75, and 85 degrees). The dataset is formatted as a set of silhouette sequences, making the data different from those in the other sets. Examples of silhouettes from this database are shown in the third and fourth columns of Fig.~\ref{fig3}.
We applied our algorithm to the OU-ISIR data to determine if silhouette masks are sufficient for gait recognition. As the algorithm we use for pose estimation requires full images, we did not create body part patches and extracted only full body features.
We used a subset of the OU-ISIR database comprising two walks taken from 1,912 subjects to meet the protocols of benchmarks \cite{cross_view}.

\subsection{Performance evaluation}\label{Performance Evaluation}

All the experiments were conducted in the following manner. The feature extractor was trained on a training set containing all data for approximately one-half of all subjects in the database (155 for TUM-GAID, 64 for CASIA, and 956 for OU-ISIR). The rest of the subjects were used for fitting the final classifier and testing the overall algorithm. The fitting components comprised four normal walks per person, while the testing components contained a further six walks (including two pairs of walks under different additional conditions). We randomly sampled 64 training subjects from the CASIA base, while the splits for the TUM-GAID and OU-ISIR datasets were provided by their authors.

For each of the testing videos, the algorithm returns the vector of probability distribution over all subjects from the testing set. We evaluated the quality of classification computing \textit{Rank-1} and the \textit{Rank-5} metrics that defined the ratio of videos in which the correct label was among the top five classifiers answers. We also plotted cumulative match characteristics (CMC) curve to compare different techniques more clearly.

Additionally to identification experiments we compared the verification quality of different methods. All the videos for testing subjects were divided into training and testing parts the same way as in the identification task. For each pair of training and testing videos the algorithm returns the distance between the corresponding feature vectors and predicts if there is the same person on both video sequences according to this distance. To evaluate the quality of the verification we plotted receiver operating characteristic (ROC) curves of false acceptance rates (FAR) and false rejection rates (FRR), and calculated equal error rates (EERs).

\subsection{Experiments and results}\label{Experiments and results}

The goal of all experiments was to explore the influence of different conditions on gait performance, including:
\begin{itemize}
\item network architecture and aggregation methods;
\item joints used for training and testing the algorithms;
\item length of captured walk.
\end{itemize}

In addition, we investigated the generality of the algorithms by training them on one dataset and applying them to another, as we expected that algorithms that depend only on body motion would work equally well on different databases.

We compared our results with those produced by two similar neural approaches \cite{me,castro} and one approach based on Fisher vectors \cite{pfm} which have demonstrated state-of-the-art results.

The first set of experiments aimed at evaluating the approach itself and comparing different technical methods such as network architectures, aggregation methods, and similarity measures. 
All the algorithms were trained from scratch. The results are shown in Table~\ref{tab3}, from which it is seen that our approach was most accurate and that it outperformed all the state-of-the-art methods.
\begin{table}[h]
	\centering
	\caption{Results on TUM-GAID dataset.\label{tab3}}{
		\begin{tabular}{|p{0.65\linewidth}|c|c|}\hline
			 {Method} & \multicolumn{2}{|c|}{Evaluation} \\\hline
			 Architecture, Aggregation and Metrics  & Rank-1 & Rank-5 \\\hline
			 VGG (PCA 1000), avg, $L_2$ & 96,44 & 100,00\\
			 VGG (PCA 1000), avg, $L_1$ & 97,84 & 100,00\\
			 VGG (PCA 500), concat, $L_2$ & 97,41 & 99,89\\
			 VGG (PCA 500), concat, $L_1$ & 98,81 & \textbf{100,00}\\\hline
			 Wide ResNet (PCA 230), avg, $L_2$ & 98,27 & 99,89\\
			 Wide ResNet (PCA 230), avg, $L_1$ & 99,24 & 99,89\\
			 Wide ResNet (PCA 150), concat, $L_2$ & 98,81 & 99,78\\
			 Wide ResNet (PCA 150), concat, $L_1$ & \textbf{99,78} & 99,89\\\Xhline{4\arrayrulewidth}
			 
			 VGG+blocks, $L_1$ \cite{me}  & 97,52 & 99,89\\\hline
			 CNN+SVM \cite{castro}, $L_2$ & 98,00 & 99,60\\\hline
			 PFM \cite{pfm} &99,20&99,50\\\hline
		\end{tabular}}{}
\end{table}

In Table 3, 'Avg' denotes a naive approach for feature aggregation in which  the mean descriptor is computed over all body parts, while 'concat' defines the concatenation of descriptors, which produces better results.

It is worth noting that in the PFM \cite{pfm} approach the input frames had initial size $640\times480$ and the resolution was not changed. Even though we reduced the inputs, the quality of the algorithms was not only maintained but even improved. Although the Rank-5 metric produced using the VGG-like network was larger, the difference was very small and the WideResNet architecture required many fewer parameters, suggesting that the  latter architecture is more appropriate for solving the gait recognition problem.

It is also interesting to note that the $L_1$ metric always produced higher accuracy of classification, suggesting that it is more suitable for measuring similarity of gait feature vectors.

\begin{figure}
    \centering
    \includegraphics[width = 1.\linewidth]{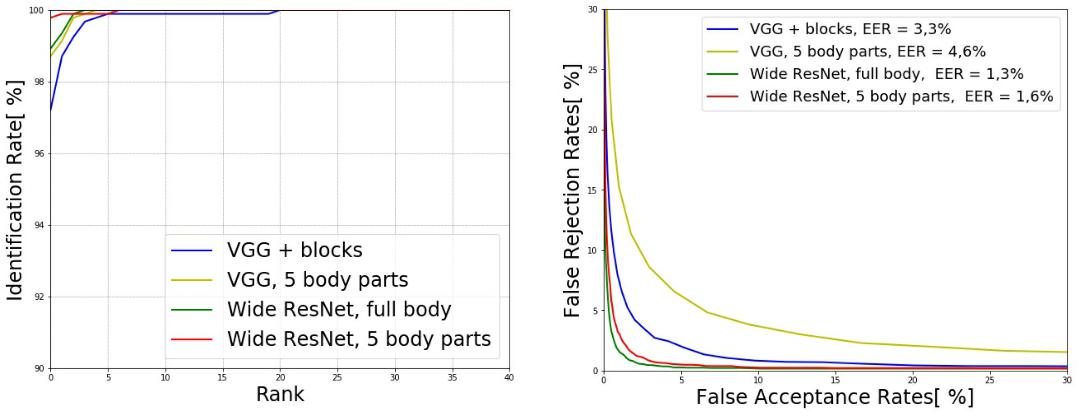}
    \caption{CMC (left) and ROC (right) curves for TUM-GAID dataset under different settings.}
    \label{tum_curves}
\end{figure}

To compare the identification and verification qualities of different settings graphically we plotted CMC and ROC curves and calculated EER. These metrics are shown in Fig.~\ref{tum_curves}.

As the Wide ResNet architecture was dominantly successful in the evaluations, all further experiments were conducted using this network structure.

Table~\ref{tab4} shows a comparison of the performance on the CASIA dataset to the results using blocks of OF maps \cite{me}. The accuracy is quite high despite the use of the rather poor training set; this might be attributable to the relatively small number of parameters in the Wide ResNet architecture and the corresponding lack of overfitting that could appear in the previous model \cite{me}.
\begin{table}[h]
	\centering
	\caption{Results on CASIA dataset.\label{tab4}}{
		\begin{tabular}{|l|c|}\hline
			 Architecture, Aggregation and Metrics  & Rank-1 \\\hline
			 Wide ResNet (PCA 150), avg, $L_2$ & 85,11 \\
			 Wide ResNet (PCA 130), avg, $L_1$ & 86,68 \\
			 Wide ResNet (PCA 170), concat, $L_2$ & 84,85 \\
			 Wide ResNet (PCA 170), concat, $L_1$ & \textbf{92,95} \\\hline
			 
			 VGG + blocks \cite{me}  & 74,93 \\\hline
		\end{tabular}}{}
\end{table}

In the experiments on the OU-ISIR database, we trained the network using all the maps and all the view angles for each training subject. During testing, we looked at gallery and probe views: as the gallery views, we used those fixed at 85 degrees, while the probe views comprised all other angles. The NN classifier was fitted onto the gallery frames from the first walk and tested on the probe frames in the second walk. 

We compare our method with several deep and non-deep approaches: view transformation models wQVTM \cite{wQVTM} and  TCM+ \cite{TCM}, methods LDA \cite{lda} and MvDA \cite{cross_view} based on discriminant analysis, bayesian approaches \cite{deep_gait_jb,bayes}, and GEINet \cite{geinet}.
The results and comparisons are shown in Table~\ref{tab5} (Rank-1) and Table~\ref{tab5_eer} (EERs).

\begin{table}[h]
	\centering
	\caption{Comparison of the classification accuracy on OU-ISIR dataset obtained from silhouette masks.\label{tab5}}{
		\begin{tabular}{|l|c|c|c|}\hline
			 {Method} & \multicolumn{3}{|c|}{Rank-1 } \\\Xhline{4\arrayrulewidth}
			 \backslashbox{Model}{Probe View}  & 55 & 65 & 75 \\\hline
			 Wide ResNet, $L_1$ &92,1 & 96,5 & 97,8  \\\hline
			 Wide ResNet, $L_2$, & & &  \\
			 extended training set &95,9& 97,8& 98,5  \\\Xhline{4\arrayrulewidth}
			 GEINet \cite{geinet} & 81,4 & 91,2 & 94,6 \\\hline
			 LDA \cite{lda} & 87,5 & 96,2 & 97,5 \\\hline
			 MvDA \cite{cross_view} & 88,0 & 96,0 & 97,0 \\\hline
			 wQVTM \cite{wQVTM} & 51,1 & 68,5 & 79,0 \\\hline
			 TCM+ \cite{TCM} & 53,7 & 73,0 & 79,4 \\\hline
            
			 {Joint Bayesian (JB)} \cite{bayes}& \textbf{94,9} & \textbf{97,6} & \textbf{98,6} \\\hline
			 DeepGait + JB \cite{deep_gait_jb}  & 89,3 & 96,4 & 98,3 \\\hline
			 \end{tabular}}{}
\end{table}

\begin{table}[h]
	\centering
	\caption{Comparison of EERs on OU-ISIR dataset obtained from silhouette masks.\label{tab5_eer}}{
		\begin{tabular}{|l|c|c|c|}\hline
			 {Method}  & \multicolumn{3}{|c|}{EER } \\\Xhline{4\arrayrulewidth}
			 \backslashbox{Model}{Probe View}   & 55 & 65 & 75\\\hline
			 Wide ResNet, $L_1$& \textbf{0,9}&\textbf{0,8} &\textbf{0,8} \\\hline
			 Wide ResNet, $L_2$,  & & & \\
			 extended training set  &0,6 &0,5 &0,5 \\\Xhline{4\arrayrulewidth}
			 GEINet \cite{geinet} & 2,4 & 1,6 & 1,2\\\hline
			 LDA \cite{lda} & 6,1 &3,1& 2,3\\\hline
			 MvDA \cite{cross_view} & 6,1 & 4,6 & 4,0\\\hline
			 wQVTM \cite{wQVTM} & 6,5 & 4,9 & 3,7\\\hline
			 TCM+ \cite{TCM} & 5,5 & 4,4 & 3,7\\\hline

			 {Joint Bayesian (JB)} \cite{bayes}&  2,2 & 1,6 & 1,3\\\hline
			 DeepGait+JB \cite{deep_gait_jb} & 1,6 & 0,9 & 0,9\\\hline
			 \end{tabular}}{}
\end{table}

The experimental results revealed that the proposed algorithm can be generalized to multiview and can obtain high accuracy even in cases of partial information. 
The second row in Tables~\ref{tab5},\ref{tab5_eer} shows the results of the network trained on all the available subjects except 956 testing ones. Such extended training set containing 2888 subjects allowed us to improve the quality and outperform the state-of-the-art method for some view angles.

\begin{figure*}[ht]
    \centering
    \includegraphics[width = 1\linewidth]{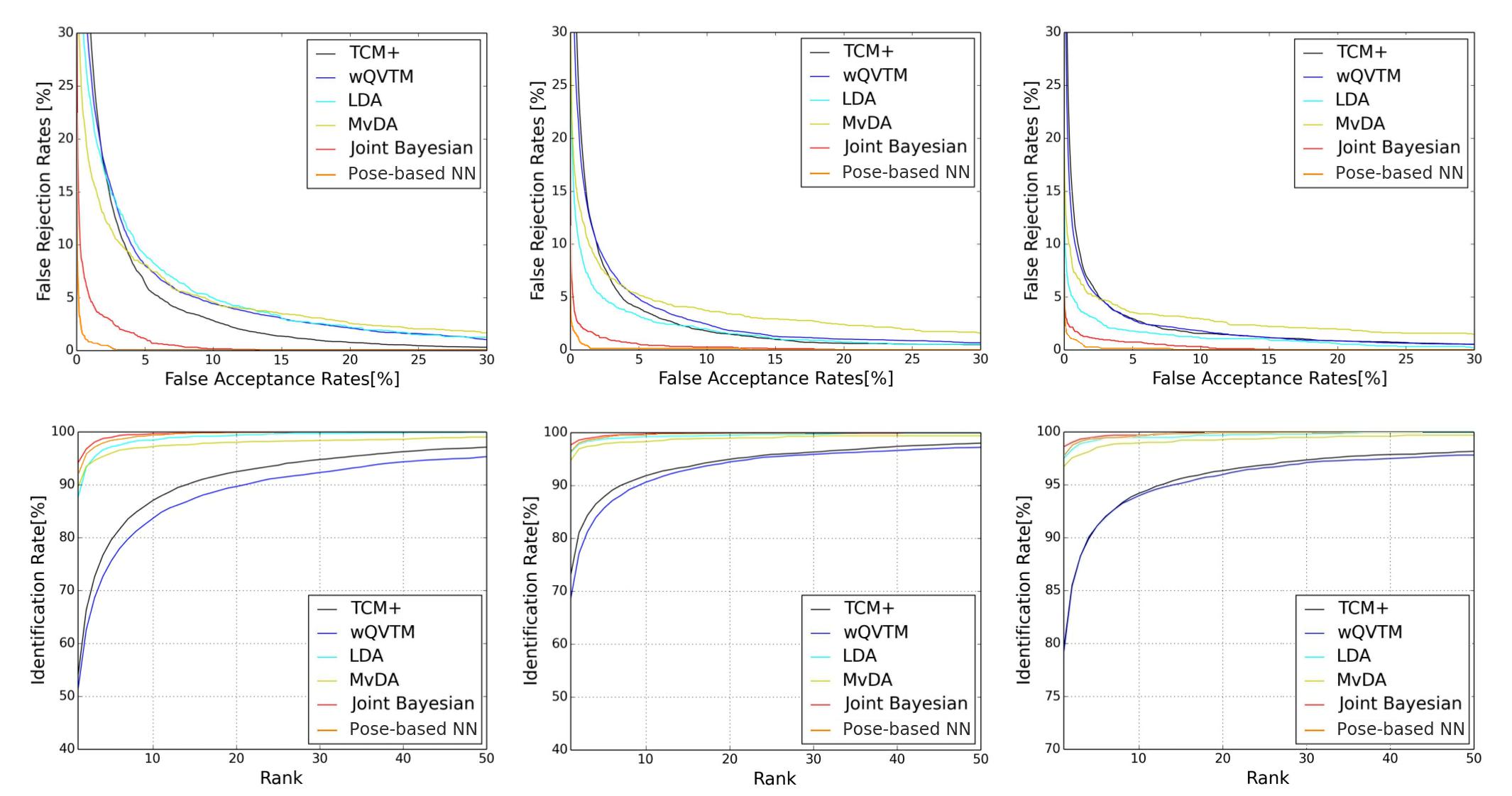}
    \caption{ROC (the first row) and CMC (the second row) curves under $85\degree$ gallery view and $55\degree$, $65\degree$, and $75\degree$ probe view, respectively.}
    \label{roc_isir}
\end{figure*}

Fig.~\ref{roc_isir} shows the ROC and CMC curves under the fixed gallery view 85 degrees. The curves of other methods were provided by their authors.
Although the recognition accuracy of our method is a bit lower than the best one \cite{bayes}, the quality of verification turns out to be higher. 

After the experiments using different training techniques, we investigated which body parts are most important in gait recognition. Results using the OU-ISIR database revealed that full-body features are quite informative; therefore, we compared different sets of body parts. We trained the network in three modes: on all five parts (each leg, upper and lower body, and full body); on three lower parts (each leg and lower body); and, finally, using only the full body as input. The results for TUM-GAID are shown in Table~\ref{tab6}.
\begin{table}[h]
	\centering
	\caption{Comparison of the results on TUM-GAID dataset obtained using different parts of the body.\label{tab6}}
		{\begin{tabular}{|l|c|c|}\hline
			 
			 Body parts  & Rank-1 & Rank-5\\\hline
			 legs, lower body, upper body, full body & \textbf{99,78} & 99,89\\
			 legs, lower body & 96,22 & 98,70\\
			 full body & 98,92 & 100,00\\\hline
		\end{tabular}}{}
\end{table}

It is seen from the table that the legs were the least informative body parts, with the network trained on the lower parts of the body producing the worst results. Instead, the algorithm built on the full body demonstrated the highest accuracy. 

Our third avenue of interest was determining the length of video needed for good gait recognition. In all of the preceding experiments, we used the entire video sequence, totaling up to 90 frames per sequence. Table~\ref{tab7} lists the results of testing the algorithms on shortened sections of TUM-GAID sequences.
\begin{table}[h]
	\centering
	\caption{Comparison of results for different lengths of videos.\label{tab7}}{
		\begin{tabular}{|c|c|c|}\hline
			 {Length of video} & {Rank-1 } & {Rank-5 } \\\hline
			 50 frames & 94,28 & 94,93\\
			 60 frames & 97,52 & 97,84 \\
			 70 frames & 99,35 & 99,46 \\
			 full length & 99,78 & 99,89 \\\hline
		\end{tabular}}{}
\end{table}

Although the length of the gait cycle is about 1 s, or $30-35$ frames, such short sequences were found to be insufficient for good individual recognition. Increasing the number of consecutive frames for classification improved the results. The expanded time of analysis was required because body point motions are similar but not identical for each step and, correspondingly, using long sequences makes recognition more stable to small inter-step changes in walking style. It is worth noting that video sequences may consist of a non-integer number of gait cycles but still be accurate recognized.

In a final set of experiments, we examined the stability and transferability of the proposed algorithm. If the feature extractor is actually general and does not depend on the background or the appearance of a person, it should be able to extract features from videos even if they were recorded under conditions different from the initial video. To check such generality, we used both the available datasets in our experiment by attempting to train the algorithm on one of the databases and evaluating its quality on the other without fine-tuning. One database was used to train the neural network and to get feature extractor and the final classifier was fitted and tested on the second one.
Table~\ref{tab8} shows the accuracy of transference of the algorithm between datasets.

\begin{table}[h]
	\centering
	\caption{Quality of transfer learning.\label{tab8}}{
		\begin{tabular}{|c|c|c|}\hline
			 \backslashbox{Training Set}{Testing Set} & CASIA & TUM \\\hline
			 CASIA & 92,68 & 66,48\\\hline
			 TUM & 76,76 & 99,78\\\hline              
		\end{tabular}}{}
	
\end{table}

Note that, even when the algorithm was trained on the other dataset, it worked on CASIA dataset B better than the method \cite{me} trained on CASIA. Nevertheless, the accuracy of recognition deteriorated considerably following transference between databases, with the error increasing by a factor of three. With TUM-GAID, the results are even worse. Training the classifier on CASIA resulted in recognition of only $66,5\%$ of the TUM testing videos, which is a factor of $1,5$ worse than that obtained by the algorithm trained on TUM. 
This suggests that the algorithm overfits and that the amount of training data (particularly in the CASIA dataset) is not sufficient for constructing a general algorithm.

\section{Implementation details}\label{Implementation details}
Some of the auxiliary methods were implemented using public libraries. The bounding boxes for human figures were computed using silhouette masks found by background subtraction. This is quite a rough method, but as each frame in the databases contained only one moving person, it worked well. The maps of optical flow were calculated by applying the Farneback \cite{Farneback} algorithm to the OpenCV library. The poses were evaluated using the heatmap regression \cite{bulat} to find the key points of the body. For the main part of the algorithm, we used Lasagne with a Theano backend and trained the networks on an NVIDIA GTX 1070 GPU. The main wide residual network employing five body parts was trained on the TUM-GAID dataset in $10$ hours.  The model was optimized using the Nesterov Momentum gradient descent method with the learning rate reduced from 0.1 by a factor of 10 each time the training quality stopped increasing.

\section{Conclusions and further work}\label{Conclusions and further work}
In this paper, we proposed a pose-based convolutional neural model for gait recognition. Our experimental results demonstrated that, although sufficiently high accuracy can be obtained only by using optical flow maps for the full-height region, collecting additional information from regions around the joints improves the results, surpassing the state-of-the-art on TUM-GAID. Our model can also be successfully applied to the moving silhouettes in OU-ISIR, which shows that the most important information for gait recognition is the movement of external edges, and that our method can be straightforwardly applied to multiview gait recognition.

\section{Acknowledgment}\label{Acknowledgment}
This work was supported by grant RFBR \#16-29-09612 "Research and development of person identification methods based on gait, gestures and body build in video surveillance data".

\bibliographystyle{ieee}
\bibliography{refs}

\end{document}